\documentclass{article}
\usepackage[nonatbib, final]{neurips_2023_ml4ps}
\usepackage[pdftex]{graphicx} 
\usepackage[utf8]{inputenc} 
\usepackage[T1]{fontenc}    
\usepackage{hyperref}       
\usepackage{url}            
\usepackage{booktabs}       
\usepackage{amsfonts}       
\usepackage{nicefrac}       
\usepackage{microtype}      
\usepackage[dvipsnames, table]{xcolor}

\usepackage[
    backend=biber,
    style=numeric-comp,
    sorting=none,
    maxcitenames=1,
    maxbibnames=2,
]{biblatex}

\DeclareFieldFormat*{citetitle}{\enquote{#1}}
\title{Robust Ocean Subgrid-Scale Parameterizations \\ Using Fourier Neural Operators}

\author{%
    Victor Mangeleer\\
    University of Liège\\
    \texttt{vmangeleer@uliege.be}\\
    \And Gilles Louppe\\
    University of Liège\\
    \texttt{g.louppe@uliege.be}\\
}

\definecolor{blue_p}{HTML}{797ef6}
\definecolor{orange_p}{HTML}{fca30b}
\definecolor{green_p}{HTML}{208a1f}
\definecolor{red_p}{HTML}{c20800}
\definecolor{grey_p}{HTML}{b2abab}
\definecolor{citation_blue}{HTML}{6d6afc}

\begin{document}
\maketitle

\begin{abstract}
In climate simulations, small-scale processes shape ocean dynamics but remain computationally expensive to resolve directly. For this reason, their contributions are commonly approximated using empirical parameterizations, which lead to significant errors in long-term projections. In this work, we develop parameterizations based on Fourier Neural Operators, showcasing their accuracy and generalizability in comparison to other approaches. Finally, we discuss the potential and limitations of neural networks operating in the frequency domain, paving the way for future investigation.
\end{abstract}

\section{Introduction}

\paragraph{Ocean simulations} Conducting ocean simulations has become increasingly important, particularly to understand and fight climate change. Despite growing computing power, a five-year Earth-scale simulation of our oceans at a 1 km resolution still takes a month worth of computation \cite{challengecomputation}. One way to address this issue is by reducing the simulation resolution to 15 km which makes it run 25 times faster. However, this approach comes at the expense of neglecting a significant portion of subgrid-scale processes contributions, i.e., the (sub)mesoscale (1-10 km) physical phenomena occurring at a scale below the simulation resolution, which highly influence the ocean dynamics.

\paragraph{Parameterizations} The challenge of achieving both speed and accuracy in ocean simulations has been approached through the development of parameterizations. The idea is to conduct a low-resolution simulation to take advantage of the fast computational time and correct it at each timestep by adding the missing subgrid-scale contributions using approximate parameterizations. Their development in simulation applications such as turbulent flows \cite{ClosureAnalytical3, ClosureAnalytical4} and, in our case, geophysical flows \cite{ClosureAnalytical1, ClosureAnalytical2, ClosureAnalytical51, ClosureAnalytical52} has been a major research effort. Nevertheless, these parameterizations all encounter the same challenges: the closure problem with extra equations, the restricted working range and the introduction of an artificial viscosity, which disturbs the turbulent nature of the flow. In response to these challenges, the exploration of deep learning-based parameterizations started. For instance, Fully-Convolutional Neural Networks \cite{FCNN1} show spatially accurate dynamics correction but have generalization and interpretability limitations. On the other hand, Fourier Neural Operators \cite{FNO} (FNO) have demonstrated notable improvements in terms of generalization and scale invariance \cite{FFNO_fourcastnet, SFNO, FFNO_unet}. To address the challenge of generalization, attention pivoted towards model discovery techniques \cite{ClosureDataDrivenZanna} to extract analytical parameterizations from raw data even if their results often lag behind neural networks \cite{Benchmarking}, indicating the need for improvements. 

\paragraph{Contributions} We show the effectiveness of Fourier Neural Operators-based (FNO) parameterizations in addressing generalization issues and accurately correcting simulations. Furthermore, we discuss extensively their advantages and limitations compared to well-established geophysical-flow and deep learning-based parameterizations. All code, datasets, results, and analysis are publicly available at \url{https://github.com/VikVador/Ocean_FNO}

\newpage

\section{Background}

\paragraph{Ocean Flow Dynamics} We focus on developing parameterizations for mid-latitude ocean flows simulations. In this region, the ocean flow dynamics can be simulated using the quasi-geostrophic equations of motion \cite{BookQG}. At each step of the simulation, all flow variables (e.g. velocity fields) are retrieved by solving the prognostic equation for the potential vorticity field $q$. In a numerical simulation, the actual equation solved is the averaged (denoted \ $\bar{\cdot}$\ ) prognostic equation given by
\begin{equation}
\frac{\partial \bar{q}}{\partial t}+(\mathbf{\bar{u}} \cdot \nabla) \bar{q} = \bar{\mathcal{F}} + \bar{\mathcal{D}} + \bar{\mathcal{S}},
\label{EQ - Potential vorticity prognostic equation}
\end{equation}
where $\mathbf{u}$ represents the velocity field vector, $\mathcal{F}$ denotes the forcing terms and $\mathcal{D}$ stands for the dissipation terms. The neglected subgrid-scale processes contributions are taken into account by adding the quantity $\bar{\mathcal{S}}$ at each simulation timestep. Our objective is to develop an accurate parameterization that quickly approximates $\bar{\mathcal{S}}$ to eliminate energy deficiencies.

\paragraph{Fourier Neural Operators} FNOs are neural network architectures that represent and process data in the frequency domain through Fourier transforms.  Depending on the problem physics, the frequency domain representation is more suited to extract intricate input-output relationships, such as demonstrated in the case of fluid mechanics where FNOs have shown strong performance \cite{FFNO_fourcastnet, FFNO_fourcastnet2, FFNO_unet}. An interesting property of this representation is that space-time convolution becomes a simple multiplication in the frequency domain, which leads to faster convolution operations.
The FNO architecture faces limitations when scaled to deep architectures, leading to the development of the Factorized Fourier Neural Operator (FFNO) \cite{FFNO}. By rethinking the architecture and enabling shared use of convolution matrices at different computation points, the FFNO achieves higher computational speed, reduces trainable parameters, and facilitates more extensive scaling.

\section{Experimental Setup}

\paragraph{Data} In mid-latitude regions, ocean flows are categorized into two types: \textit{jet-driven flow}, characterized by predominant motion in a single direction, and \textit{eddy-driven flow}, marked by chaotic swirling motion. To simulate these ocean dynamics, we use the two-layer quasigeostrophic model from PyQG \cite{pyqg}. This model offers the necessary complexity to faithfully represent both types of flow. Specifically, the upper layer simulates interactions with the atmosphere, while the lower layer represents interactions with the ocean bottom. It is important to note that these layers also interact with each other. For training, the parameterization takes as inputs ($u$, $v$, $q$) $\in \mathbb{R}^{H \times H}$ where $u$ and $v$ are the ocean flow longitudinal and latitudinal velocity fields, and $H$ the grid resolution. The parameterization output is the missing subgrid-scale contributions denoted as $\bar{\mathcal{S}}$. The loss function is the mean squared error between the ground truth and the value predicted by the parameterization.

\paragraph{Protocol} We assess the quality of our parameterizations through a standardized benchmarking framework \cite{Benchmarking} and compare against several baselines. These baselines include a Fully-Convolutional Neural Networks (FCNN) \cite{FCNN1}, U-NET \cite{unet}, and common empirical geophysical parameterizations such as Smagorinsky \cite{ClosureAnalytical2}, Backscattering and Biharmonic \cite{ClosureAnalytical51}, and one derived through symbolic regression on raw data \cite{ClosureDataDrivenZanna, Benchmarking}. For the neural networks, they use the same inputs, predict identical quantities, and each consists of approximately 300,000 trainable parameters. Furthermore, the FCNN is used in its optimal configuration presented in \cite{Benchmarking}, and the hyperparameters of the U-NET have been adjusted to make its capacity match the one of the others. These two models will not undergo any further fine-tuning. The primary objective is to assess, at a fixed capacity, whether a model exhibits superior performance compared to others. Afterwards, the model demonstrating the most promising capabilities, in this case FFNO, will undergo further tuning for optimization. We conducted two tests: offline and online. \textbf{Offline} testing means evaluating how effectively the parameterization predicts $\bar{\mathcal{S}}$ on a dataset composed of 5000 samples drawn from ten distinct, previously unobserved jet-driven simulations. For each fluid layer, we compute the coefficient of determination $R^2$, yielding a value of 1 for perfect predictions, 0 when predictions are no better than guessing the mean, and negative values when predictions are worse than guessing the mean. \textbf{Online} testing means evaluating the parameterization performance at making long-term projections. This involves analyzing whether the parameterized low-resolution simulation dynamics align better with those in the corresponding high-resolution simulation, in contrast to the low-resolution simulation.

\newpage 

To do so, we examine the power spectrum of energy-related flow properties. If the observed spectrum closely matches the one from the high-resolution simulation, it confirms the parameterization effectiveness in redistributing and addressing energy deficiencies across various scales during the simulation.

\section{Results}

\paragraph{Offline} The results are summarized in Tab.\ref{TAB - Offline Results}. In the first row, we establish a baseline using the training conditions outlined in \cite{Benchmarking}, which used a dataset of 5000 samples from a single 10-year-long eddy-driven simulation. This allowed us to evaluate both accuracy and generalization across different flow types. Several key observations emerge: both FCNN and U-NET yield negative $R^2$ results in both layers. FNOs exhibit a positive score in the upper layer. Only FFNO achieves a positive score in both layers. In summary, the FNOs demonstrate superior generalization abilities. In the second row, we investigate the impact of mixing different eddy-driven simulations while maintaining a constant total number of samples. This led to improvements in all $R^2$ values. In the third row, we increased the number of samples. Surprisingly, we observe a decrease in overall $R^2$ values, with one exception: FFNO. This suggests that all models may have overfit the eddy dynamics during training. However, we speculate that \textbf{the FNOs ability to work with spectral data representations allows it to extract more complex interactions relationships between large-scale and subgrid-scale processes thus leading to better generalization performance}. In the fourth row of results, we explore the impact of training on both ocean flow dynamics. FCNN and U-NET models achieve positive $R^2$ values in the upper layer but struggle with negative values in the lower layer. In contrast, both FNOs consistently achieve positive $R^2$ values, highlighting once again the effectiveness of spectral representation computations. As our last experiment, we recognized FFNO as a promising parameterization and we proceeded to fine-tune it across various configurations, with detailed results summarized in the appendix. This led to the development of FFNO*, the best configuration obtained with a substantially larger capacity, i.e. nearly 20 times larger than the one of FFNO. For this last experiment, we observe that \textbf{FFNO* achieves unparalleled accuracy, sparking the possibility of extracting a more complex and refined analytical expression for the parameterization through symbolic regression}.

\paragraph{Online} As shown in Fig.\ref{FIG - Online results}, while the FFNO (\textcolor{green_p}{green}) demonstrated good generalization abilities (see Tab.\ref{TAB - Offline Results}, third row), it failed to reproduce the high-resolution (\textcolor{blue_p}{blue}) simulation spectral profiles. This discrepancy arises from its limitations in accurately predicting the lower layer dynamics, which significantly impacts the long-term simulation and once again highlight the importance of finding good parameterizations. In contrast, FFNO* (\textcolor{red_p}{red}) displayed superior and consistent results across all flow variables (except APEflux) compared to the baselines (\textcolor{grey_p}{grey}). Indeed, the baseline models exhibited inconsistencies, either causing spectral explosion or flattening. Despite the fact that FFNO* achieves superior results with comparable capacity to other models, it faces the drawback of using twenty-four Fourier layers, as opposed to the original FNO. Consequently, \textbf{the speed-up due to utilizing a deep-learning-based parameterization for low-resolution simulations is compromised}. This is shown in Tab.\ref{TAB - Computational time results}, where the increased computational time, attributed to the frequent computation of Fourier transform and inverse Fourier transform, surpasses the one of the high-resolution simulation.

\textbf{In conclusion, leveraging the frequency representation of data allows Fourier Neural Operators to exhibit superior generalization by capturing multiscale interactions. Nevertheless, there is still a need to enhance the FFNO architecture. This improvement is crucial not only for refining parameterization generalization capabilities but also for reducing simulation computational time, ultimately making the development of Fourier Neural Operator-based parameterizations worthwhile}.

\newpage 

\section{Conclusion}
Our exploration of Fourier Neural Operators (FNOs) as subgrid-scale parameterizations demonstrates their superior ability to generalize when predicting missing subgrid-scale contributions in new ocean dynamics. However, further refinement is needed for the learned parameterization since its predictions lack the necessary accuracy for long-term projections. When learning a parameterization across all possible ocean dynamics, FNOs again outperform the baselines in terms of generalization. This reinforces the notion that FNOs are better suited for this task. Moreover, after optimization, FNOs achieved remarkable accuracy levels. Yet, we must acknowledge that the computational time for the parameterized low-resolution model is three times longer than that of the high-resolution counterpart thus making it unusable. Furthermore, while the parameterization effectively corrected energy deficiencies, spatial accuracy remained a concern. Even if the predictions were accurate, their incorrect placement had a noticeable impact on flow dynamics. 

\bgroup
\def\arraystretch{1.6}
\begin{table}[!b]
  \caption{Analysis of our parameterizations accuracy by calculating the coefficient of determination ($R^2$) for both ocean model fluid layers. A $R^2$ value close to 1 indicates better results. The \textit{Training Datasets} column provides information on training sample sizes and the number of simulations used. Testing is performed using a dataset of jets-driven flow simulations. The key observations are: FFNO generalizes better in every situations (row 1 to 3). When trained on both ocean dynamics, FCNN and U-NET perform poorly on the lower layer whereas the FNOs obtained positive score suggesting again better generalization capabilities. When the training and architecture are optimized, FFNO* reaches new levels of accuracy suggesting that extracting the analytical expression of the learned parameterization is a path worth to explore.}
  \label{TAB - Offline Results}
  \resizebox{\columnwidth}{!}{
  \begin{tabular}{cccccccc}
    \toprule[1.7pt]
    \multicolumn{3}{c}{Training Datasets} & \multicolumn{5}{c}{Upper layer $R^2$\ /\ Lower layer $R^2$} \\
    \cmidrule[1pt](r){1-3}\cmidrule[1pt](r){4-8}
    Samples [$10^3$]  & Eddies & Jets & FCNN & U-NET & FNO & FFNO & FFNO* \\
    \cmidrule[0.5pt](r){0-0}\cmidrule[0.5pt](r){1-1}\cmidrule[0.5pt](r){2-2}\cmidrule[0.5pt](r){3-3}\cmidrule[0.5pt](r){4-4}\cmidrule[0.5pt](r){5-5}\cmidrule[0.5pt](r){6-6}\cmidrule[0.5pt](r){7-7}\cmidrule[0.5pt](r){8-8}
    5  & 1  & 0 & -0.21 \ /\ -10.60 & -0.69 \ /\ -16.30 & 0.18 \ /\ -0.92 & 0.76 \ /\ 0.12                    & -  \\ 
    5  & 10 & 0 & -0.23 \ /\ -11.80 & -0.40 \ /\ -12.34 & 0.42 \ /\ -0.23 & 0.78 \ /\ 0.24                    & -  \\ 
    20 & 40 & 0 & -0.34 \ /\ -15.20 & -0.43 \ /\ -14.40 & 0.41 \ /\ -0.34 & \textcolor{green_p}{\textbf{0.86}} \ /\  \textcolor{green_p}{\textbf{0.37}} & -  \\ 
    \bottomrule[0.1pt]
    5  & 5  & 5 &  0.40 \ /\ -5.75  & 0.32  \ /\ -4.91  & 0.49 \ /\  0.13 & 0.83 \ /\ 0.49                    & \textcolor{green_p}{\textbf{0.99}} \ /\  \textcolor{green_p}{\textbf{0.93}}  \\
    \bottomrule[1.7pt]
  \end{tabular}}
\end{table}
\bgroup

\begin{table}[!b]
  \caption{Analysis of the computational time for a 10-year jet-driven flow scenario simulation reveals the following: The first two columns display the baseline times for high and low-resolution simulations, while the next four columns show the total time needed for a parameterized low-resolution simulation. The speed-up metric quantifies the improvement in computational time compared to a high-resolution simulation. A speed-up above 1 indicates a positive gain—meaning the simulation is faster, with higher values being better—while a speed-up below 1 signifies a negative gain, indicating slower performance than the high-resolution simulation. Key findings indicate that only the FCNN, U-NET, and FNO parameterizations yield positive speed-ups, despite their worst predictive accuracy. On the other hand, while FFNO and FFNO* exhibit superior correction capabilities, their simulation times surpass that of a high-resolution simulation, negating the advantage of using parameterizations. This discrepancy is attributed to the FFNO's twenty-four Fourier layers compared to the FNO, resulting in significant computational overhead due to the numerous Fourier transformations and inverse Fourier transforms required between layers.}
  \label{TAB - Computational time results}
  \centering
  \def\arraystretch{1.8}
  \resizebox{\columnwidth}{!}{
    \begin{tabular}{cccccccc}
      \toprule[1.7pt]
      \multicolumn{1}{c}{} & \multicolumn{2}{c}{Baselines [min]} & \multicolumn{5}{c}{Parameterized low-resolution (LR) [min]} \\
      \cmidrule[1pt](r){2-3} \cmidrule[1pt](r){4-8}
      & High-resolution & Low-resolution & FCNN & U-NET & FNO & FFNO & FFNO* \\
      \cmidrule[0.5pt](r){2-2}\cmidrule[0.5pt](r){3-3}\cmidrule[0.5pt](r){4-4}\cmidrule[0.5pt](r){5-5}\cmidrule[0.5pt](r){6-6}\cmidrule[0.5pt](r){7-7}\cmidrule[0.5pt](r){8-8}
      & 22.25  & 2.10 & 11.09 & 13.05 & 11.397 & 46.89 & 61.68  \\ 
      \bottomrule[1pt]
      Speed-up [-] & 1.00 & 10.58 & 2.00 & 1.70 & \textcolor{green_p}{\textbf{1.95}} & \textcolor{red_p}{\textbf{0.47}} & \textcolor{red_p}{\textbf{0.36}}  \\
      \bottomrule[1.7pt]
    \end{tabular}
  }
\end{table}

\newpage

\section{Future work}
For future work, we propose investigating novel architectures based on Fourier Neural Operators. To address spatial challenges, we suggest exploring a hybrid approach by combining space-time convolution with spectral convolution. In terms of computational efficiency, there is potential for improvement in the design of Fourier layers to reduce their number. Finally, we find it intriguing to explore the application of symbolic regression to extract learned parameterizations.

\acksection
This work was  supported by Service Public de Wallonie Recherche under Grant No. 2010235 - ARIAC by DIGITALWALLONIA4.AI. In addition to that, the authors would like to thank François Rozet, Gerome Andry, Omer Rochman and Sacha Lewin for their support and insightful feedback.

\begin{figure}[!t]
    \includegraphics[width=1\linewidth, trim={0 0 9.2cm 0}, clip]{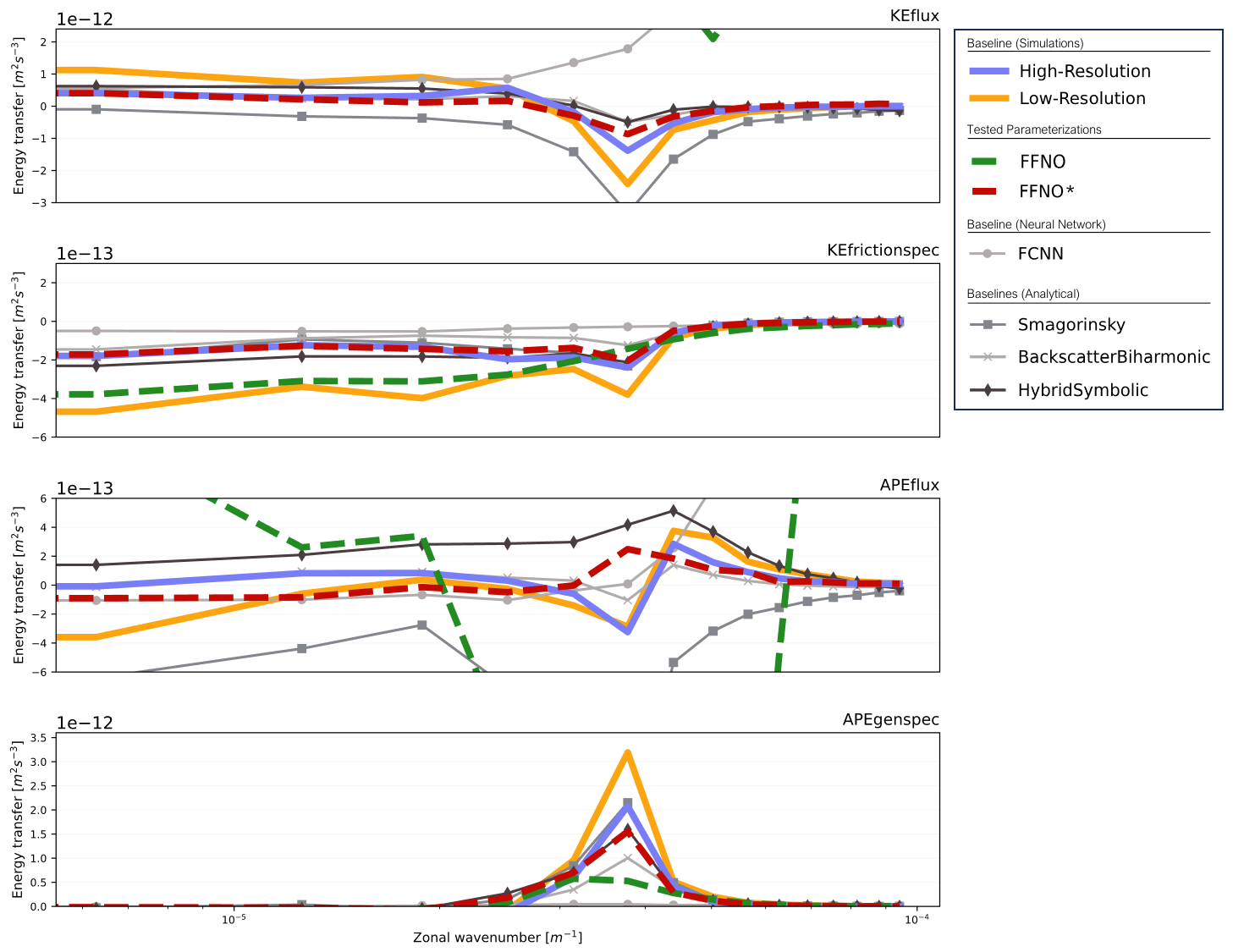}
    \caption{Analysis of energy transfer contributions (y-axis) across various length scales (x-axis) for different flow quantities. The objective is to align the power spectrum with that of the high-resolution (\textcolor{blue_p}{blue}). The observed flow quantities are: \textit{KEflux} is the kinetic energy transfer, while \textit{KEfrictionspec} assesses energy dissipation due to friction with the bottom of the ocean. \textit{APEflux} measures available potential energy for transfer, and \textit{APEgenspec} tracks newly generated potential energy at each scale. Observations reveal that while FFNO (\textcolor{green_p}{green}) exhibits good offline generalization (see Tab.\ref{TAB - Offline Results}, third row), its predictions lack the accuracy necessary for a stable long-term simulation, resulting in spectra divergence from the high-resolution. Conversely, the optimized FFNO* (\textcolor{red_p}{red}), trained on both flow dynamics types, demonstrates improved matching with the high-resolution spectra (except for APEflux). Baselines (\textcolor{grey_p}{grey}) do not achieve the same consistency as FFNO*, they either excessively amplify the spectra, worsening them compared to the low-resolution (\textcolor{orange_p}{orange}), or significantly dampen them (indicated by the flat lines for KEfrictionspec).}
    \label{FIG - Online results}
\end{figure}

\newpage

\section*{References}
\printbibliography[heading=none]
\newpage

\section{Appendix}
\begin{figure}[!b]
    \centering
    \includegraphics[width=0.9\linewidth, trim={0 0 0cm 0}, clip]{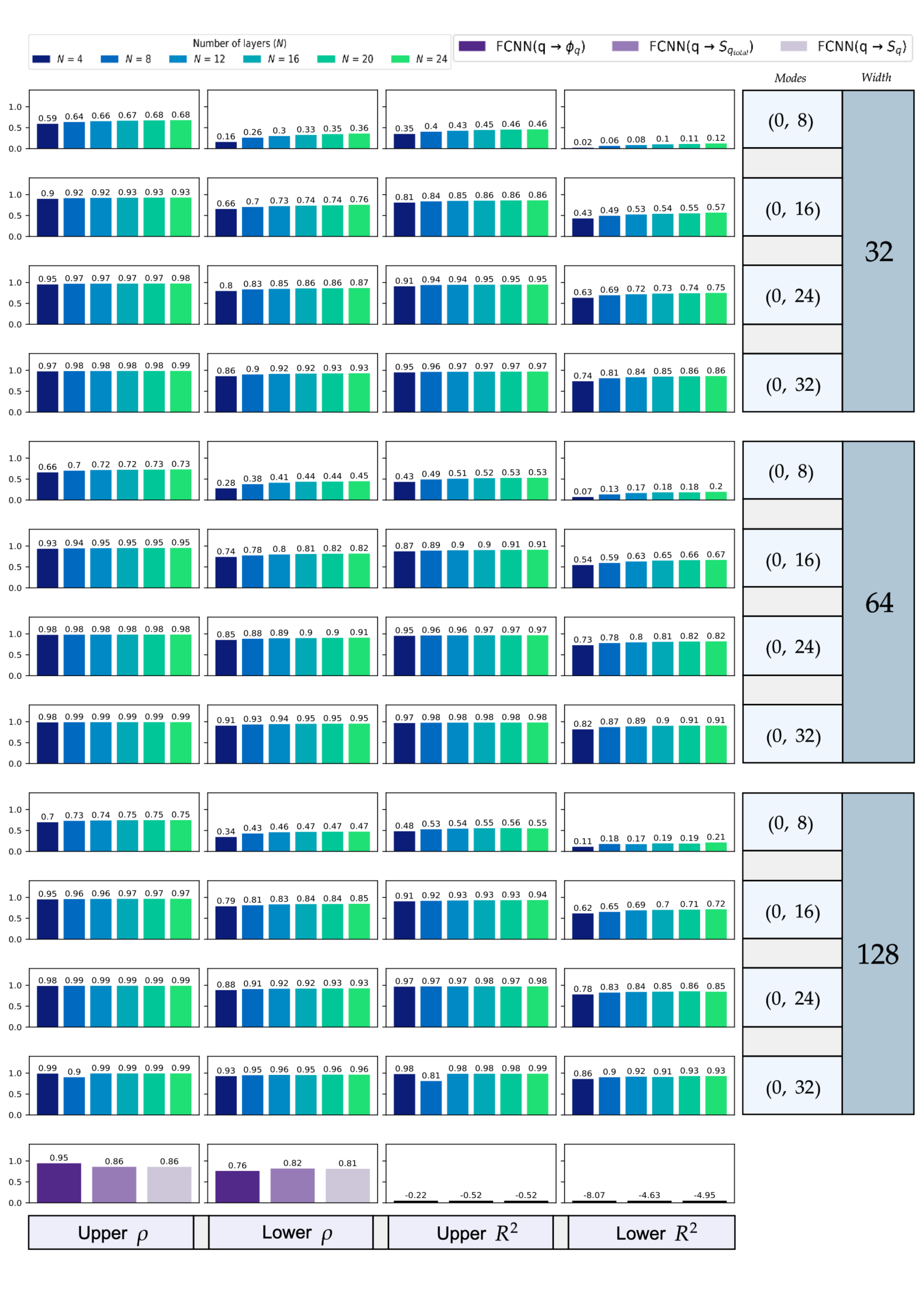}
    \caption{Sensitivity analysis of FFNO hyperparameters on jet-driven flow. Models trained on a dataset with eddies and jet-driven flows are evaluated offline. The first two columns depict correlations in the upper and lower layers, while the next two show $R^2$ coefficients. Hyperparameters include \textit{width} (latent space size), the number of Fourier layers \textit{N}, and the range of Fourier modes. Notably, the number of retained modes appears to have the most impact, with broader ranges yielding better results. Increased Fourier layers generally show a minor impact but significantly prolong the forward pass time due to additional Fourier transform computations. Enhanced \textit{width} improves model capacity and results, with a lesser impact on computational time compared to an increase in depth. As a result, FFNO uses a width of 128, the first 32 modes, and 24 Fourier layers.}
    \label{FIG - Online results}
\end{figure}
\end{document}